\documentclass{article}
\usepackage{IJCAI-2}  
\usepackage{epsfig}
\usepackage[cmex10]{amsmath}
\usepackage{subfigure}
\usepackage[ruled,linesnumbered]{algorithm2e}
\usepackage{multirow}
\usepackage{diagbox}
\usepackage{flushend}
\usepackage{bbm}
\usepackage{xcolor}
\usepackage{subfigure}
\usepackage[export]{adjustbox}
\usepackage[hyphens]{url}  
\usepackage{graphicx} 
\usepackage{graphicx}  
\title{Region-Wise Attack: On Efficient Generation of Robust Physical Adversarial Examples}

\author{{\small
    Bo Luo and Qiang Xu\\
    Department of Computer Science \& Engineering \\
    The Chinese University of Hong Kong, Shatin, N.T., Hong Kong\\
   Email: \{boluo,qxu\}@cse.cuhk.edu.hk}
}

\begin{document}
\maketitle

\begin{abstract}
Deep neural networks (DNNs) are shown to be susceptible to adversarial example attacks. Most existing works achieve this malicious objective by crafting subtle pixel-wise perturbations, and they are difficult to launch in the physical world due to inevitable transformations (e.g., different photographic distances and angles).
Recently, there are a few research works on generating physical adversarial examples, but they generally require the details of the model a priori, which is often impractical.
In this work, we propose a novel physical adversarial attack for arbitrary black-box DNN models, namely \emph{Region-Wise Attack}. To be specific, we present how to efficiently search for region-wise perturbations to the inputs and determine their shapes, locations and colors via both top-down and bottom-up techniques. In addition, we introduce two fine-tuning techniques to further improve the robustness of our attack. 
Experimental results demonstrate the efficacy and robustness of the proposed \emph{Region-Wise Attack} in real world.


\end{abstract}

\section{Introduction}
Deep neural networks (DNNs) have achieved the state-of-the-art performance in many areas, such as image classification~\cite{he2016deep,zoph2018learning} and autonomous driving~\cite{chen2016monocular,yang2018end}. However, they are found vulnerable to adversarial example attacks that fool them into making adversarial decisions by slightly manipulating their inputs~\cite{papernot2016limitations,carlini2017towards,xie2019improving,yuan2019adversarial}, which are serious threats to safety-critical systems.

Instead of attacking the digital inputs to the DNNs, physical adversarial attacks manipulate the objects in real world directly to achieve malicious objectives. The first such kind of attack was proposed in~\cite{sharif2016accessorize}, in which attackers wear a malicious eye-glasses to fool the face recognition system to make misclassifications. In~\cite{eykholt2018robust}, a physical adversarial attack was implemented against road sign recognition systems by generating sticker perturbations and attaching them onto traffic signs. However, these attacks are launched under white-box settings, in which attackers need to know the details of the attacked DNN model (e.g., architectures and trained parameters). This is often not practical due to the difficulty to obtain the parameters used in real world systems.

In this work, we propose to generate physical adversarial examples by searching for effective perturbations in continuous image regions with simple queries of the targeted DNN model without knowing its details. The proposed attack, namely \emph{Region-Wise Attack}, is able to efficiently (i.e., with a limited number of queries) determine the locations, shapes and colors of the required perturbations to launch the attack.
To further increase the robustness of \emph{Region-Wise Attack}, physical misplacement and photography independent fine-tuning mechanisms are introduced to tolerate possible variations in real world. The physical attack can be launched by sticking the generated perturbations onto targeted objects. Experimental results on CIFAR-10, GTSRB data sets and real world road signs demonstrate the efficacy and robustness of the proposed attack.

\vspace{2pt}

The main contributions of this paper include:
\begin{itemize}


\item We propose a novel physical adversarial attack for arbitrary black-box DNNs by generating region-wise perturbations.

\item We present how to efficiently find the shapes, locations and colors of the region-wise perturbations via both top-down and bottom-up methods.

\item To increase the attack robustness, we introduce two fine-tuning mechanisms to tolerate possible physical misplacement and photographic variations in real world.


\end{itemize}


\section{Related Work and Motivation}\label{sec:rw}


\begin{figure*}[ht]
\centering
\includegraphics[width=0.9\textwidth]{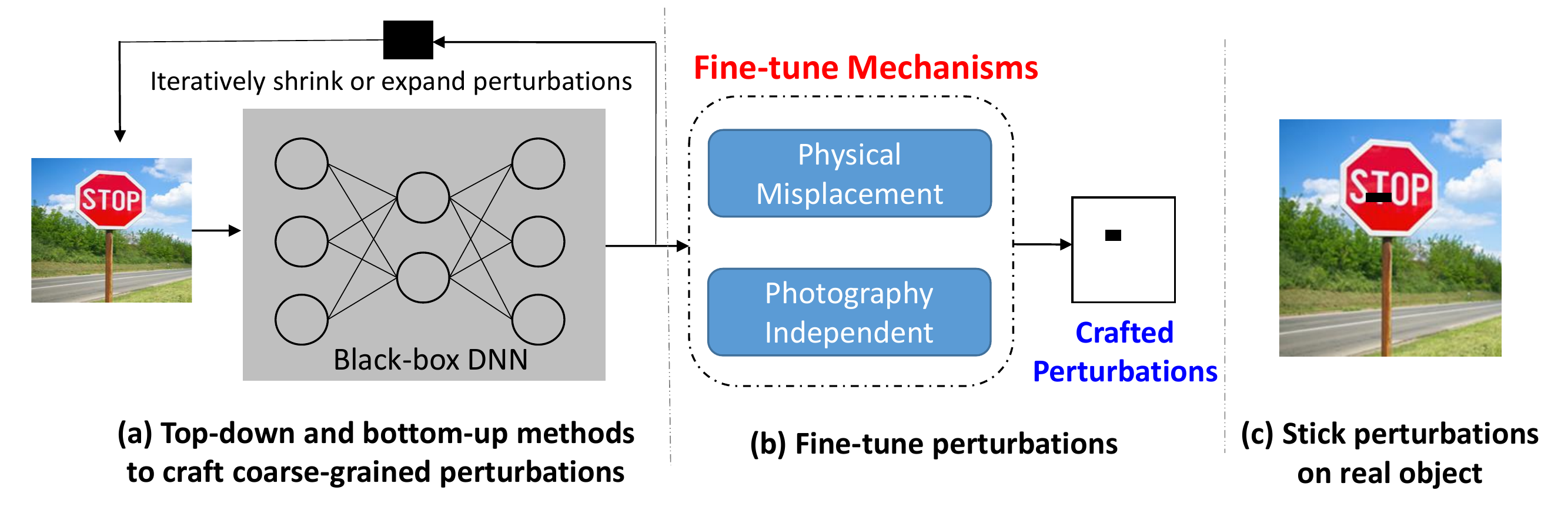}
\vspace{-5pt}
\caption{The overview of our proposed attack method. 
}\label{fig:overflow}
\end{figure*}

In many machine learning systems, attackers can only manipulate the objects so that the taken images are perturbed as expected. This kind of attacks is called physical attacks. \cite{sharif2016accessorize} targets to attack the face recognition systems by generating adversarial eye-glasses. When people wear them, the system will make misclassifications. \cite{eykholt2018robust} generates adversarial stickers to mimic city graffiti and then attach these stickers onto road signs to fool self-driving systems. In~\cite{thys2019fooling}, they propose to craft adversarial patches to mislead the object detection systems. Recently, ~\cite{komkov2019advhat} generates rectangular perturbations on the hat to fool face ID systems.

However, all the above physical attacks are under white-box settings, which is often not practical, as it is difficult to obtain the DNN parameters of real-world systems. There are also many related black-box attacks proposed in the literature. In~\cite{papernot2017practical}, they first train a substitute network of the similar functionality with the targeted DNN, and then craft adversarial examples against the substitute network under white-box settings. The generated adversarial examples are used to attack the targeted DNN based on their transferability. The ZOO attack~\cite{chen2017zoo} crafts adversarial examples by approximating the model gradients, observing the output changes when varying the inputs. They use zeroth order optimization method to approximate the model gradients rather than training a substitute model. Recently,~\cite{alzantot2018genattack} proposes the Genattack to find the adversarial perturbations with genetic algorithms. These attacks craft pixel-wise perturbations and generally require extensive computational resources to achieve better transferability. More importantly, they are less effective for launching physical attacks due to inevitable physical misplacement and photographic variations in real world. 

Motivated by the above, we propose novel techniques to launch robust physical adversarial example attacks for arbitrary black-box DNN systems by efficiently crafting region-wise perturbations, in which we treat the targeted DNN as an oracle and only query the outputs for specific inputs. 

\section{Crafting Region-wise Perturbations}
\label{sec:method}

Crafting region-wise perturbations under black-box settings is very difficult, as we have to determine the perturbation color, shape and location. The search space is extremely large. The perturbation color refers to how to change the values of the selected regions. For example, we can set the selected perturbation region of the input image to the black color and then attach this black stick onto the object to launch the attack. The perturbations with different colors, shapes or locations will certainly result in different attack effects.

The overview of our proposed attack method is shown in Figure~\ref{fig:overflow}, in which we first generate coarse-grained region-wise perturbations with our top-down or bottom-up methods iteratively under black-box settings. Then, to improve the attack robustness in the physical world, we discuss two fine-tuning techniques to tolerate possible variations. After that, the crafted region-wise perturbations are attached onto the physical objects so that the taken images can mislead the attacked DNNs. The key steps in our method are the top-down, bottom-up techniques and fine-tuning mechanisms, as detailed in the following sections. 

\subsection{Top-down Method for Un-targeted Attacks}
For un-targeted adversarial example attacks, the attack goal is to fool the model into making misclassifications. The simplest way to achieve this objective is to perturb the whole object region in the image so that the model cannot recognize it. 
However, this is not valid, as perturbing the whole object region will certainly raise human suspicions.
Therefore, to successfully launch the attack, we need to shrink the perturbation region to reduce human attentions.
This is the key idea of our top-down method that shrinks the perturbation regions iteratively until it cannot mislead the targeted classifier. 

In our top-down method, firstly, we will set the perturbation as a specific color and the perturbation region is initialized as the whole input image. 
Secondly, we shrink the perturbation region into some smaller candidate regions. Thirdly, we select the most adversarial ones from the candidate regions by querying the attacked DNN. The second and third steps are repeated until the constraint on the size of the perturbation region is satisfied. If the selected color cannot attack successfully under the constraint, then we will use another color to attack. 
The key steps in this process is how to shrink the perturbation regions and how to select the most adversarial ones, which are detailed in the following subsections.

\noindent \textbf{Region Shrinking.}
In each iteration, we will generate some smaller candidate perturbations based on the current region-wise perturbation, and evaluate their individual attack capability by querying the targeted DNN. Some principles should be considered in the region shrinking process. Firstly, the union of all the generated smaller candidate regions should cover all parts of the original region. If not, some parts in the original region will not be explored and we may miss the optimal regions. Secondly, we cannot generate too many candidates, otherwise, the number of querying the targeted DNN will be large.

\begin{figure}[!htb]
\centering
\includegraphics[width=0.9\columnwidth]{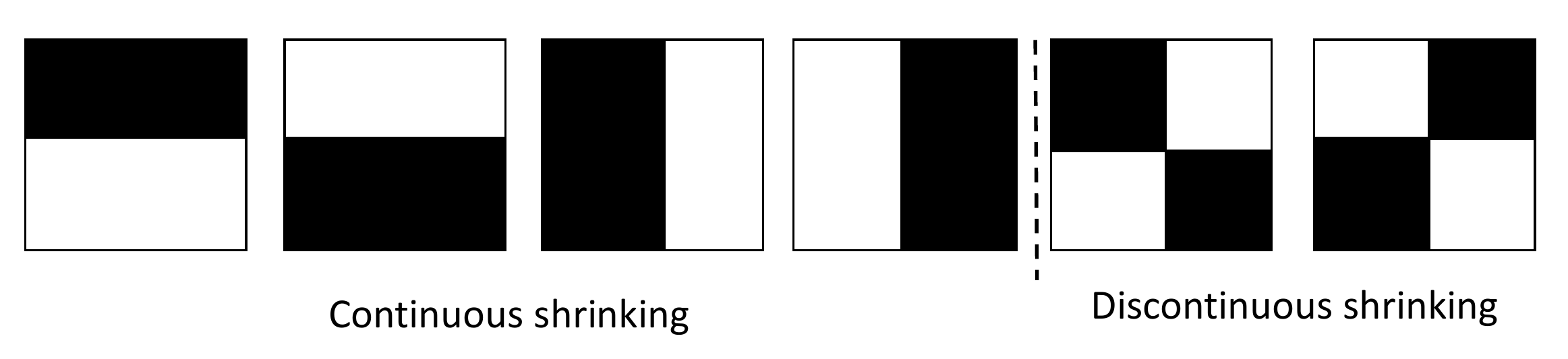}
\caption{The continuous and discontinuous shrinking.
}\label{fig:top_down_shrinkage}
\end{figure}

Considering these two principles, we propose to divide the original perturbation region horizontally and vertically in each iteration to get four different continuous candidate regions, the left-half, right-half, top-half and bottom half parts, respectively. 
To increase the attack success rate, we also generate discontinuous candidate regions, the top-left and bottom-right parts or the top-right and bottom-left parts, respectively. In each iteration, the region will shrink to its half size, and no parts in the original region are missed. Figure~\ref{fig:top_down_shrinkage} shows an example for the continuous and discontinuous region shrink strategies, in which the black region will be perturbed.


\noindent \textbf{Region Selection.}
After generating the candidate perturbation regions, we need to evaluate their attack capability and select the most adversarial ones as the search space for the next iteration.
According to the goal of the un-targeted attack, the region that achieves the minimum prediction probability for the true label should be selected. 
It can be formulated as follows:
\begin{equation}
r_{i+1} = \mathop{\arg\min}_{ r_i\in R}\ f_{true}(x+r_i),\\
\end{equation}
where $r_i$ is a candidate region-wise perturbation in iteration $i$, $x+r_i$ is the image with perturbation $r_i$. 

It should be noted that for each attack, we will first initialize the perturbation color, then we optimize the region shape and location. If we explore more colors, we will have a higher chance to attack. 
However, exploring more color candidates will certainly increase the attack cost and it is necessary to obtain a good tradeoff between the attack cost and performance. 

\subsection{Bottom-up method for Targeted Attacks}
Although the top-down method is effective for un-targeted attacks, it is ineffective for the targeted attacks. The reason is that large perturbation regions can only fool DNNs into making mistakes but may not necessarily make them misclassify to the targeted malicious label. As a result, the top-down method that searches from the large regions to smaller ones is not efficient in this case. To solve this problem, we propose a bottom-up technique that starts from small perturbation regions and then extends them iteratively based on their adversarial capabilities.

In the bottom-up method, firstly, we initialize the color of the perturbation, then search the initial effective perturbation regions among the whole input image with a fine-grained manner. Secondly, we expand the fine-grained perturbation regions to different directions, and generate several candidate regions. Thirdly, we evaluate these expanded regions based on their attack capabilities and select the most adversarial ones for the next iteration. The second and third steps are continued until the constraint on the size of perturbation region is violated. Next, we introduce these three steps in detail.

\noindent \textbf{Perturbation Region Initialization.}
The goal of the perturbation region initialization is to find small appropriate regions in the input images as initial locations for perturbations. We select $k$ initial perturbation regions that achieve the top $k$ prediction probabilities of the targeted label. If $k=1$, we only select the region with the largest prediction probability of the targeted label and the perturbation is continuous. However, the important features for achieving the attack goal may locate in separate regions. To increase the attack capability, it is necessary to select several initial regions to generate discontinuous region-wise perturbations, which will increase the overhead for the attacks. As a result, attackers can choose the large $k$ whenever possible based on their attack cost budgets.

We assume that the initial perturbation region is with the size of $n*n$. The exhaustive search is used to find the most adversarial regions, where the stride size is $j$ ($j \le n$). 
For example, if size of the original search space is $9*9$ and the initial perturbation region is $3*3$ with the stride 3, we have to evaluate 9 different perturbation regions, and any two search regions have no overlaps. 

\noindent \textbf{Region Expansion.}
In each iteration, we will expand the perturbation region similar to the shrinking step in the top-down method. We expand the current perturbation regions along four directions (top, bottom, left and right directions). After each iteration, we expand the original region into its twice size and get four candidate regions for the next region selection step.

\noindent \textbf{Region Selection.}
In each iteration, the region that achieves the maximum prediction probability of the targeted label is selected. It can be formulated as follows:
\begin{equation}\label{eq:bottom_up}
r_{i+1} = \mathop{\arg\max}_{\boldsymbol r_i\in R}\ f_{targeted}(x+r_i),\\
\end{equation}
where $x+r_i$ is the image with region-wise perturbation $r_i$. $f_{targeted}(x+r_i)$ denotes the classification probability of the targeted label given the perturbed input.

It should be noted that the bottom-up method can be used for un-targeted attacks, but the top-down method can achieve good attack success rate with less queries. 
So in this paper, we only adopt it for targeted attacks.

\subsection{Explore More Region-wise Shapes}
We can efficiently find the adversarial region to attack by region-shrinking or region-expansion with the proposed two methods, but the limitation is that the shape of the region-wise perturbation is always rectangular. To mitigate this limitation, we propose to explore more shapes after finding the adversarial regions. The idea is to search other shapes (such as the triangle, circle, octagonal, etc) from the sizes that they are exactly internally bounded by the generated rectangle, then iteratively increase their sizes until they can externally bound the rectangle. In this way, we can explore more region-wise shapes to launch the physical adversarial attack.

\section{Physical Fine-tuning Mechanisms}\label{sec:fine-tune}
After generating the region-wise perturbations, we need to launch the physical attack by sticking them onto the real objects. As in the physical world, there are many unavoidable variations, the adversarial examples generated from input images under certain photography conditions may fail to attack after sticking. In this section, we introduce two fine-tuning techniques to increase the attack robustness in the physical world.



\subsection{Physical Misplacement Fine-tuning}

In practical situations, it is very difficult, if not impossible, to precisely stick the perturbations onto the real objects as we optimized. For example, it may translate several pixels. As a result, to improve the attack robustness, it is necessary to fine-tune the perturbation locations to tolerate possible physical misplacement.


Our key idea is to explore the neighborhood of the generated perturbations to find the best location that can tolerate the majority of the physical misplacement. If we move the perturbation $m$ pixels every time within the neighborhood, for each $m$, we have at most eight directions, left, right, top, bottom and top-left, bottom-left, top-right, bottom-right, for physical misplacement. If the perturbation is still adversarial when suffering from the physical misplacement, it means that the perturbation can tolerate this misplacement. The more times a perturbation can tolerate the misplacement, the more robustness it is. Based on this idea, we fine-tune the perturbation to tolerate the possible physical misplacement according to the following equation:

\begin{align}
\boldsymbol m, \boldsymbol d &= \mathop{\arg\max}_{m \in M, d \in D}\ \frac {|\sum_{x \in X} (Att(x+r_{m,d})=Suc)|} {|X|},
\end{align}
where $r_{m,d}$ is the generated perturbation $r$ moving $m$ pixels along the direction $d$. $M$ is the set of values that $m$ can choose in the neighborhood of $r$. $D$ is the eight directions that $r$ can move to. The numerator denotes the number of successful attacks for all the input in $X$  when $r$ suffering from the physical misplacement. The denominator is the total number of images in $X$. According to this equation, $r$ will achieve the best misplacement tolerance when moving $\boldsymbol m$ pixels to the direction $\boldsymbol d$, which is the final perturbation location after fine-tuning.

\subsection{Photography Independent Fine-tuning}

In the physical world, many inevitable photographic variations, such as photographic angles, distances and so on, may degrade the attack robustness, as the taken images will be different due to these variations. In this section, we propose a photography independent fine-tuning mechanism to improve the attack robustness. The key idea is to generate the input images under varying photography conditions and craft the perturbation that can attack successfully under most conditions. To achieve this goal, we modify the region selection standard in the top-down and bottom-up methods. Instead of evaluating the adversarial capability of a region on one input image, we evaluate it with a number of inputs. In this way, the crafted region-wise perturbations can largely increase the attack robustness in the physical world.


The new region selection standard for the un-targeted attacks is reformulated as follows:


\begin{equation}
r_{i+1} = \mathop{\arg\min}_{ r_i\in R}\ \frac {\sum_{x\in X} f_{true}(x+r_i)} {|X|},\\
\end{equation}
where $x$ is an image of the object under one specific situation and $X$ is a set of images taken in varying conditions. We select $r_i$ that can largely degrade the average of the classification probability of the true label for all the inputs in $X$.

Similarly, for the targeted attacks, the perturbation selection standard is reformulated as follows:


\begin{equation}
r_{i+1} = \mathop{\arg\max}_{\boldsymbol r_i\in R}\ \frac {\sum_{x\in X} f_{target}(x+r_i)} {|X|}.\\
\end{equation}
The perturbation that can largely increase the average of the classification probability of the targeted malicious label for all the inputs in $X$ will be selected.




In our method, the input image set $X$ is obtained by the data augmentation through physical and synthetic transformations. For example, if we target to attack the stop sign, we will generate $X$ by taking photos of the stop sign under various conditions, such as changing photographic angles, lights or object distances. 
For the synthetic transformation, we perform common digital image transformations, such as changing the contrast or rotating some angles.

\begin{figure*}[!t]
\centering
\begin{minipage}[t]{0.48\textwidth}
\centering
\includegraphics[width=0.98\columnwidth]{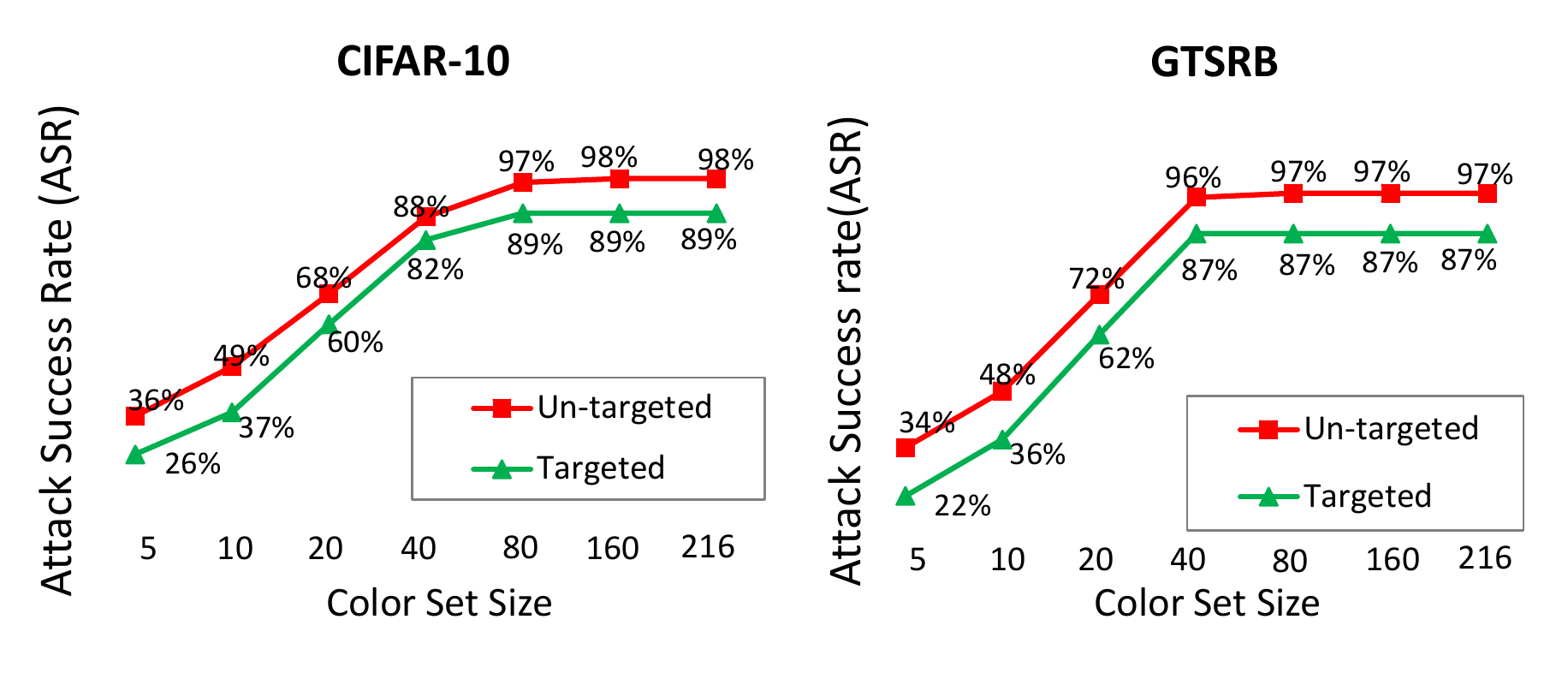}
\caption{ASR vs. color set size for un-targeted attack.}\label{fig:asr-cs}
\end{minipage}
\begin{minipage}[t]{0.48\textwidth}
\centering
\includegraphics[width=0.80\columnwidth]{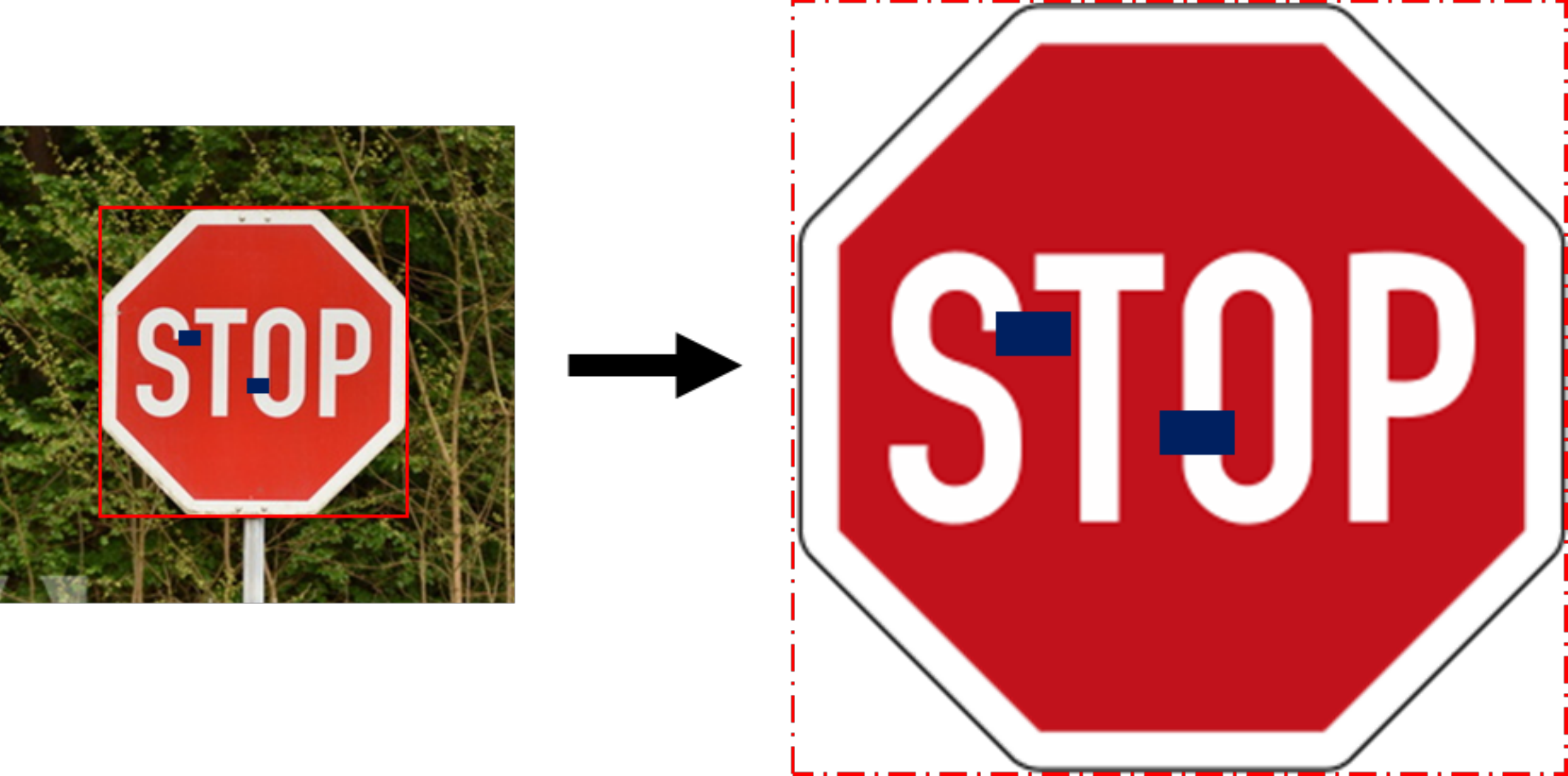}
\caption{Scale digital perturbation to physical world.}
\label{fig:pa}
\end{minipage}
\end{figure*}

\begin{table*}[!t]
\centering
\scalebox{0.8}[0.8]{
\setlength{\arrayrulewidth}{1pt}
\begin{tabular}{|cc|ccc|ccc|ccc|ccc|}
 \hline
&&\multicolumn{3}{c|}{\textbf{White-box Method}}&\multicolumn{3}{c|}{\textbf{Exhaustive Search}}&\multicolumn{3}{c|}{\textbf{Ours (Continuous)}}&\multicolumn{3}{c|}{\textbf{Ours (Discontinuous)}}\\
\cline{3-14}
& &ASR& Runtime&Query&ASR& Runtime&Query&ASR& Runtime&Query&ASR& Runtime&Query\\
\hline

\multirow{2}{*}{CIFAR-10}&Un-targeted& 93.0\%&28.1 Secs  & -&  93.2\% &178.6 Secs  & 4,410&  92.6\% &3.2 Secs  & 432&  97.4\% &8.1 Secs  & 864\\
&Targeted& 85.4\%&42.6 Secs  & -&  83.0\% &252.4 Secs  & 12,936&  82.3\% &5.2 Secs  & 568&  89.4\% &11.3 Secs  & 1,224\\
\hline

\multirow{2}{*}{GTSRB}&Un-targeted &92.3\%&39.6 Secs& -&  92.6\% &282.8 Secs&16,507& 91.7\% &4.6 Secs&504  &  96.5\% &10.3 Secs  & 1,080\\

&Targeted& 82.9\%&57.8 Secs  & -&  82.5\% &336.8 Secs & 25,368&81.2\% &8.7 Secs & 672&  87.2\% &16.8 Secs  & 1,436  \\
\hline
\end{tabular}
}
\caption{Performance compared to the baselines considering the ASR, query number, and runtime.}
\label{tb:top-down-traversal}
\end{table*}

\section{Experiments}\label{sec:experiments}

Our digital attack experiments are performed on CIFAR-10 and GTSRB (German Traffic Sign Recognition Benchmark) data sets. The size of the image in CIFAR-10 is 32*32, and the size of the input in GTSRB is 64*64.
The classification accuracy of the trained DNN model is 92.8\% for CIFAR-10 and 98.8\% for GTSRB on the test sets, which are comparable to the state-of-art results.

We conduct all the experiments on the platform with an AMD Threadripper 1950X CPU and four NVIDIA GTX 1080 GPUs.
In our bottom-up method, the initial number of perturbation regions $k$ is set to 2 for the discontinuous method. The adversarial label for the targeted attack is randomly selected.

\subsection{ASR VS. Color Set Size}
As discussed in previous sections, the more colors we explore, the larger attack success rate (ASR) we will achieve. In this section, we evaluate the influence of the color set size on the ASR. Figure~\ref{fig:asr-cs} shows the ASR of 1000 correctly classified test images on the two data sets. The color set is randomly selected from the 216 web safe colors. 
We can see that the ASR increases quickly when the color size is small, but it will saturate when the color size is greater than 80 for CIFAR-10. That is to say, when the color size is large enough, increasing the color size may not bring ASR increase. The discovery for GTSRB is similar. The ASR will not increase when the color size is greater than 40. The number of colors needed for CIFAR-10 is larger than that for GTSRB, which can be explained that the images in CIFAR-10 are more colorful than GTSRB, thus it is necessary to explore more colors to achieve the successful attack. Please be noted that even we explore all the 216 colors, some inputs are still difficult to be attacked.

\subsection{Runtime Comparison}
\noindent \textbf{Baselines} As we generate region-wise perturbations under black-box settings and there are no such existing attacks, we use a white-box attack, \cite{eykholt2018robust}, as one of a baseline, in which region-wise perturbations are crafted to mimic city graffiti. 
It should be noted that it is not practical to attack real world systems under the white-box setting and we use it as a baseline only to show the advantages of our proposed methods.
The second baseline we choose is an exhaustive search method that searches all the possible locations of the region-wise perturbations. 
We evaluate the performance of our method compared to the baselines under the constraint that the maximum number of perturbed pixels is the same. It is 1.6\% of the total number of pixels of the input image and the region shape is rectangular. 

Table~\ref{tb:top-down-traversal} shows the number of queries, ASR and runtime of three methods, in which we can see that our method with discontinuous region separation achieves the highest ASR and the lowest runtime consumption compared to the two baselines. This is because our method explores multiple colors and discontinuous regions for attack while the white-box method can only generate black region-wise perturbation, as they set perturbation regions to 1, and the exhaustive search method can only generate continuous perturbations. 
Moreover, this benefit of runtime is more obvious for GTSRB. As the query needed for the exhaustive search method increases about 4 times for CIFAR-10, it only increases 1.2 times for our methods.
This is because for the exhaustive search method, the query needed is highly relevant to the input size. Our methods use a greedy search technique, in which the query number will not be influenced much by the input size.  

\begin{figure}[!tb]
\centering
\caption{Attack performance under varying shooting distances and angles where the targeted malicious label is "Straight Drive".}\label{fig:physical-image}
\includegraphics[width=1.\columnwidth]{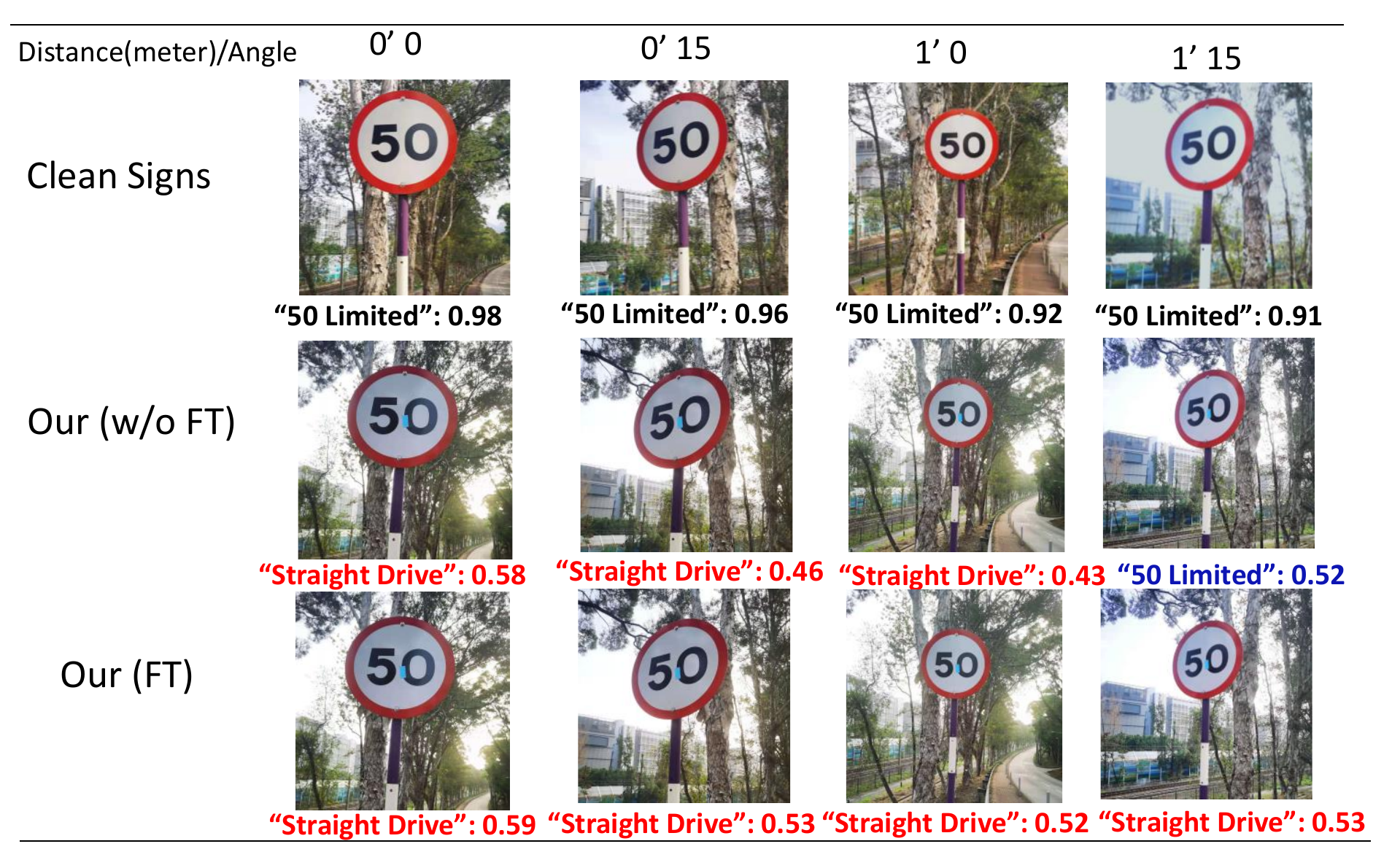}
\end{figure}

\subsection{Physical Attack Results}
\begin{table*}[!tb]
\centering
\scalebox{0.9}[0.9]{
\setlength{\arrayrulewidth}{0.7pt}
\begin{tabular}{|c|c|cc|cc|cc|}
\hline
&&\multicolumn{2}{c|}{\textbf{``50 Speed Limited"}}&\multicolumn{2}{c|}{\textbf{``Straight Drive"}}&\multicolumn{2}{c|}{\textbf{``No Entry"}} \\
\cline{3-8}
&&Targeted&Un-Targeted&Targeted&Un-Targeted&Targeted&Un-Targeted \\
\hline
\multirow{2}{*}{\textbf{Gaussian Blur}}&Our(w$/$o FT)&63.0\%& 82.6\% &64.9\%& 83.2\%&63.9\%& 82.4\%\\
&Our(FT)& 87.3\%& 95.3\%& 89.2\%& 95.8\%& 87.6\%& 94.4\%\\
\hline
\multirow{2}{*}{\textbf{Contrast Change}}&Our(w$/$o FT)&63.4\%& 82.2\% &65.3\%& 83.0\%&64.2\%& 82.6\%\\
&Our(FT)& 87.7\%& 95.4\%& 89.8\%& 95.7\%& 87.9\%& 94.9\%\\
\hline
\end{tabular}
}
\caption{Robustness of adversarial attacks under synthetic transformations. }\label{tb:fine-tune}
\end{table*}
\noindent \textbf{Physical Experimental Process.}
To evaluate the robustness of our attack in the physical world,
we conduct experiments on real world road signs. Firstly, we collect the digital images of road signs in the physical world, and then scale the images into 64*64 pixels that is the input size of the trained GTSRB model. Secondly, we send these digital images into the DNN model and craft the region-wise perturbations using the proposed attack methods. Thirdly, we scale the digital region-wise perturbations back to the real world size and stick them onto road signs.



\noindent \textbf{Physical Robustness Metric}
We compute the physical robustness (PR) of the region-wise perturbations in physical world using the following formula:
\begin{equation}
PR = \frac{\sum_{x\in X} \mathbbm{1} \{F(x+r)^T\not =y  \& F(x)^T=y\}} {\sum_{x\in X} \mathbbm{1} \{F(x)^T=y\}},\\
\end{equation}
where $r$ denotes the region-wise perturbation and $T$ denotes the transformations in real world. $y$ is the ground truth label. This formula is for un-targeted attacks. For targeted attack, the constraint $F(x+r)^T\not =y$ should change to $F(x+r)^T=y^*$ where $y^*$ is the adversarial label. This equation ensures that the misclassification is caused by the added perturbations instead of other factors.

\noindent \textbf{Digital to Physical Mapping}
After generating the digital region-wise perturbations, we should map them back to the physical world. The size of the adversarial perturbations and their locations should be scaled considering the real world traffic signs. We call this process as \emph{digital to physical mapping}, as shown in Figure \ref{fig:pa}. The idea is that using a rectangular box to bound the road signs in the digital image, and then we calculate the relative size and location of the perturbation with respect to the bounding box. Next, the perturbation for real world attacks is scaled and attached according to the size of real world road signs.


\noindent \textbf{Physical Results}
Figure~\ref{fig:physical-image} shows the physical attack performance of our region-wise perturbations under varying shooting distances and angles for targeted attacks on \emph{``50 Speed Limited''} road sign. The values below each image represent the predicted label and its prediction confidence given by the GTSRB model.
The adversarial perturbations are generated based on the digital images photographed under a specific distance and angle, denoted as 0 distance and 0 angle. We can see that our method without using the fine-tuning mechanisms can tolerate most conditions, but failed when the shooting distance and angle changed to 1 meter and 15$^\circ$. However, after using the fine-tuning mechanisms that find a more robust location near the neighborhood, the perturbation now can tolerate all the physical variations. 

To simulate other complex situations in physical world, we use synthetic transformations  to evaluate our methods. The results for contrast change and gaussian blur transformations are shown in Table~\ref{tb:fine-tune}. For each road sign, we take 100 times, and then do synthetic transformations under 10 different parameter settings. For the gaussian blur, we change the standard deviation of added noise from 0.01 to 0.1 with the step size of 0.01. For contrast change, we increase the contrast parameter from 1.1 to 2 with the step size of 0.1. Therefore, we get 1000 test samples for each transformation. From Table~\ref{tb:fine-tune}, we can see that our region-wise perturbation can achieve a good performance under different transformations, where the robustness with the fine-tuning mechanisms increases about 15\% for the un-targeted attack, and increases about 36\% for the targeted attack compared to the method without fine-tuning. These results demonstrate that our top-down and bottom-up methods with fine-tuning mechanisms are very effective to generate robust adversarial perturbations under black-box settings.

\section{Conclusion}\label{sec:conclusion}
In this paper, we propose \emph{Region-Wise Attack}, a novel technique to efficiently generate robust physical adversarial examples under the black-box setting. Top-down and bottom-up methods are presented to efficiently find the appropriate colors, locations and shapes of the region-wise perturbations. To further increase the attack robustness, we introduce two fine-tuning techniques to tolerate possible variations in the real world.  Experimental results on CIFAR-10, GTSRB and real-world attacks show that our proposed method can achieve very high attack success rate only using black-box queries.



\bibliographystyle{named}
\bibliography{ref}


\end{document}